# The Thing That We Tried Didn't Work Very Well: Deictic Representation in Reinforcement Learning

**Sarah Finney**
AI Lab
MIT
Cambridge, MA 02139

**Natalia H. Gardiol**
AI Lab
MIT
Cambridge, MA 02139

**Leslie Pack Kaelbling**
AI Lab
MIT
Cambridge, MA 02139

**Tim Oates**
Dept of Computer Science
Univ. of Maryland, BC
Baltimore, MD 21250

## Abstract

Most reinforcement learning methods operate on propositional representations of the world state. Such representations are often intractably large and generalize poorly. Using a deictic representation is believed to be a viable alternative: they promise generalization while allowing the use of existing reinforcement-learning methods. Yet, there are few experiments on learning with deictic representations reported in the literature. In this paper we explore the effectiveness of two forms of deictic representation and a naïve propositional representation in a simple blocks-world domain. We find, empirically, that the deictic representations actually worsen learning performance. We conclude with a discussion of possible causes of these results and strategies for more effective learning in domains with objects.

## 1 Introduction

Real-world domains involve objects: things like chairs, tables, cups, and people. Yet most current machine learning algorithms require the world to be represented as a vector of attributes. How should we apply our learning algorithms in domains with objects? It is likely that we will have to develop learning algorithms that use truly relational representations, as has been done generally in inductive logic programming [12], and specifically by Dzeroski et al. [5] for relational reinforcement learning. However, before moving to more complex mechanisms, it is important to establish whether, and if so, how and why, existing techniques break down in such domains. In this paper, we document our attempts to apply relatively standard reinforcement-learning techniques to an apparently relational domain.

One strategy that has been successful in the planning world [8] is to *propositionalize* what is essentially a relational domain. That is, to make an attribute vector with a single Boolean attribute for each possible instance of the properties and relations in the domain. There are some fairly serious potential problems with such a representation, including the fact that it does not give much basis for generalization over objects. Additionally, the number of bits to be considered grows exponentially with the number of objects in the world, even if the task to be accomplished does not become more complicated. An alternative to this full-propositional representation is to create a *deictic-*propositional representation that, intuitively, affords more possibility for appropriate generalization.

The word deictic was introduced into the artificial intelligence vernacular by Agre and Chapman [1], who were building on Ullman's work on visual routines [15]. A deictic expression is one that "points" to something: its meaning is relative to the agent that uses it and the context in which it is used. *The-book-that-I-am-holding* and *the-door-that-is-in-front-of-me* are examples of deictic expressions in natural language. The primary motivation for the use of deictic representation is that it avoids the arbitrary naming of objects, naturally grounding them in agent-centric terms [2]. Deictic representations have the potential to bridge the gap between relational and propositional representations, allowing much of the generalization afforded by first-order representations yet remaining amenable to solution (even in the face of uncertainty) by existing algorithms.

Our motivating intuition was that deictic representations might ameliorate the severe scaling problems of full-propositional representations. First, we can achieve passive generalization through use of the markers. For example, if I have learned what to do with *the-cup-that-I-am-holding*, it doesn't matter whether that cup is *cup3* or *cup7*. Second, since the size of the observation space in a deictic representation only grows with the number of attentional markers, our agent should be able to perform a task in domains with varying numbers of objects more easily than the full-propositional agent. Last, we expected that our deictic agent would gain an advantage from the ability to focus its attention on aspects of the world relevant to its current activity and to ignore the aspects that do not matter.

In most deictic representations, and especially those in which the agent has significant control over what it perceives, there is a substantial degree of partial observability: in exchange for focusing on a few things, we lose the ability to see the rest. As McCallum observed in his thesis [10], partial observability is a two-edged sword: it may help learning by obscuring irrelevant distinctions as well as hinder it by obscuring relevant ones.

What is missing in the literature is a systematic evaluation of the impact of switching from full-propositional to deictic representations with respect to learning performance. The next sections report on a set of experiments that begin such an exploration.

## 2 Experiment Domain

Our learning agent exists in a simulated blocks world. It must learn to pick up a green block by first removing any blocks covering it. This problem domain was introduced by Whitehead and Ballard [16] in their early work on deixis in relational domains. They developed the Lion algorithm to deal with the domain's partial observability by avoiding aliased states. McCallum [9] showed that this partial observability could be directly handled by keeping a short history of observations.

For example, if the agent is currently looking at, say, a red block in the domain shown in Figure 1, it cannot tell if it should proceed to search for the green block, or if it should pick up that red block in order to clear it from off the top of the green block. By examining the last action and observation, however, the agent knows that if it has just looked up from a green block, then it is in the state where it should clear off the red block.

The experiments described in this section differ from previous empirical work with deictic representations in two important ways. First, our goal was to compare the utility of the different representations, rather than to evaluate or develop a learning algorithm tailored for one representation. Second, we have not tuned the perceptual features, actions, or training paradigm to the task but instead developed a set of perceptual features and actions that seemed reasonable for an agent that might be given an arbitrary task in a blocks world.

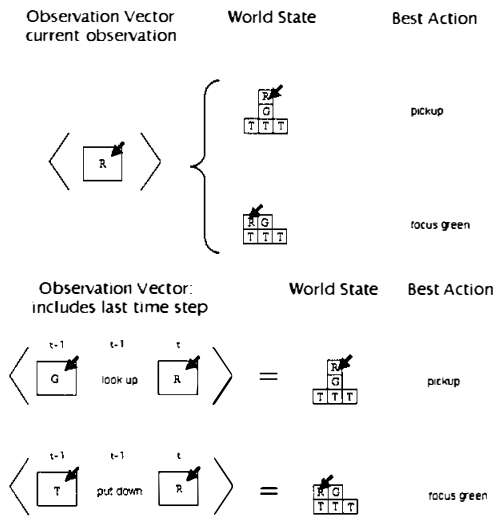

Figure 1: Adding history to handle partial observability.

### 2.1 Two Deictic Representations

A deictic name for an object can be conceived as a long string like *the-block-that-I'm-holding*, an idea that can be implemented with a set of markers. For example, if the agent is focusing on a particular block, that block becomes *the-block-that-I'm-looking-at*; if the agent then fixes a marker onto that block and moves its attention somewhere else, the block becomes *the-block-that-I-was-looking-at*.

For our experiments, we developed two flavors of deictic representation. In the first case, called "focused" deixis, there is a focus marker and one additional marker. The agent receives all perceptual information relating to the focused block: its color (**red**, **blue**, **green**, or **table**), and whether the block is in the agent's hand. In addition, the agent can identify a marker bound to any block that is above, below, left of, or right of the focus. The second case, called "wide" deixis, receives perceptual information (color and identities of any adjacent markers) for each marked block, not just the focused block. The action set for both deictic agents is:

- *move-focus(direction)*: The focus cannot be moved beyond the top of the stack or below the table. If the focus is to be moved to the side and there is no block at that height, the focus falls to the top of the stack on that side.

- *focus-on(color)*: If there is more than one block of the specified color, the focus will land randomly on one of them.

- *pick-up()*: This action succeeds if the focused block is a non-table block at the top of a stack.

- *put-down()*: Put down the block at the top of the stack being focused.

- *marker-to-focus(marker)*: Move the specified marker to coincide with the focus.
- *focus-to-marker(marker)*: Move the focus to coincide with the specified marker.

### 2.2 Full-Propositional Representation

In the fully observable propositional case, arbitrary names are assigned to each block. The agent can perceive a block's color, the location of the block, and the name of any block under it. In addition, there is a single bit that indicates whether the hand is holding a block. The propositional agent's action set is:

- *pick-up(block#)*: This action succeeds only if the block is a non-table block at the top of a stack.
- *put-down()*: Put down the block at the top of the stack under the hand.
- *move-hand(left/right)*: This action fails if the agent attempts to move the hand beyond the edge of the table.

### 2.3 Comparing State and Action Spaces

Propositional representations yield large observation spaces and full observability; deictic representations yield small observation spaces and partial observability. We examine the concrete implications of this next.

Our experiments used two different blocks-world starting configurations (Figure 2). The number of distinct[1] block arrangements is 12 in the *blocks1* setup and 60 in *blocks2*. The underlying state space in the full-propositional case depends on all the ways to name the blocks. This yields 5760 ground states in *blocks1* and 172,800 in *blocks2*. The size of the observation space, however, outpaces the number of ground states dramatically: roughly 10 billion in *blocks1* and roughly 3 trillion in *blocks2*.[2]

The underlying state space in the deictic case depends on possible marker locations rather than on block names. This gives a total of 1200 ground states in *blocks1* and 8640 in *blocks2*. The size of the observation space, however, is constant in both domains: the size of the focused deictic observation space is 512, and that of the wide deictic observation space is 4096 [6].

The action set for the deictic representations does not change with additional blocks, so it is constant at 12 actions. The full-propositional representation requires an additional *pickup()* action for each block, so it has five possible actions in *blocks1* and six in *blocks2*.

[1]Blocks of the same color are interchangeable.
[2]Note that this observation space corresponds to the size needed for a look-up table, and it includes many combinations of percepts that are not actually possible.

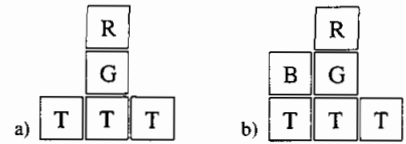

Figure 2: The two blocks-world configurations: a) *blocks1* and b) *blocks2*. The table is made of unmovable table-colored blocks.

## 3 Learning Algorithms

In these experiments, we took the approach of using model-free, value-based reinforcement learning algorithms, because it was our goal to understand their strengths and weaknesses in this domain. In the conclusions, we discuss alternative methods.

As discussed previously, because we no longer observe the whole state in the deictic representation, we have to include some history. The additional information requirement renders the observation space too large for an explicit representation of the value function, like a look-up table. Thus, we required learning algorithms that can approximate the value function.

We chose Q-learning with a neural-network function approximator (known as neuro-dynamic programming [3], or NDP) as a baseline, since it is a common and successful method for reinforcement learning in large domains with feature-based representation. We hoped to improve on the performance of NDP by using function approximators that could use history selectively, such as the G algorithm [4] and McCallum's U-Tree algorithm [10]. After some initial experiments with U-Tree, we settled on using a tree-growing algorithm based on the G algorithm and U-Tree.

### 3.1 Neuro-Dynamic Programming

Our implementation of NDP used a two-layer back-propagation neural network for each action. The input to each network was a vector containing the current observation plus some number of previous observations and actions. The input layer to each network consists of one node per possible action and observation feature value; a given input vector is encoded by setting the appropriate input nodes to one and the rest to zero. The output of each network is the estimated Q-value for that action in the state represented by the input vector. As has been observed by others [14], we found that SARSA led to more stable results than Q-learning because of the partial observability of the domain.

## 3.2 Tree-Growing Algorithm

The G algorithm uses a tree structure to determine which elements of the state space are important for predicting reward. A leaf in the tree is the agent's internal representation for a state, and it corresponds to a series of perceptual distinctions useful for predicting reward. Each leaf determines the agent's policy by estimating Q-values for the possible outgoing actions. The tree is initialized with a root node that makes no state distinctions. The root has a fringe of nodes beneath it, where each fringe node represents a possible further distinction. Statistics are kept in the root node and the fringe nodes about reward received during the agent's lifetime. A statistical test determines whether any of the distinctions in the fringe are worth adding permanently to the tree; that is, it looks for statistical differences between the reward values of a parent node and its fringe nodes. If a further distinction is found to be statistically significant, the fringe nodes under that parent become official leaves, and a new fringe set is created beneath each new leaf node.

At this level of description, the G algorithm is essentially the same as U-Tree. The major distinction between them is that U-Tree requires much less experience with the world at the price of greater computational complexity: it remembers historical data and uses it to estimate a model of the environment's transition dynamics, and then uses the model to choose a state to split. The G algorithm makes each new splitting choice based on direct estimates of the Q values from new data. U-Tree uses a non-parametric statistical test (the Kolmogorov-Smirnov test) for splitting nodes, which is more robust than the test used by the original G. The G-based algorithm used in this work extends G by using the Kolmogorov-Smirnov test and by considering distinctions over a finite history window (rather than just the current observation). See the technical report [6] for a discussion of the differences between G and U-Tree, and the way in which the G algorithm of this paper differs from the original.

## 4 Experiment Outcomes

We conducted a set of learning experiments in the blocks-world environments shown in Figure 2. The task in both cases was to pick up the green block. The agent received a reward whenever it succeeded at the task, a penalty if it took an action that failed (e.g., attempted to move its hand off the edge of the world, or attempted to pick up the table), and a smaller penalty for each step otherwise. The agent used an $\epsilon$-greedy exploration strategy, with $\epsilon = 0.10$.

The left plot in Figure 3 shows the results from running the NDP algorithm in the *blocks1* domain. The graph shows the scaled total reward[3] received in a testing trial plotted against the number of training steps: at the end of each set of 200 training steps, the state of the learning algorithm was frozen and the agent took a 100-step testing trial during which the total accumulated reward was measured; exploration was not turned off during testing. Each curve for *blocks1* is averaged over 10 experiments, and for *blocks2* over five experiments.

From the graph, we see that the deictic representations did not immediately show the edge we anticipated. We expected, then, to see them gain an advantage with the addition of a distractor block, as in *blocks2*. The results are shown on the right side of Figure 3. Rather than surpassing, or even approaching, the performance of the full-propositional agent, the deictic agents performed worse than before.

Clearly, by adding additional blocks yet retaining the same observation space, we were aggravating the partial observability for the deictic agents. Since selectively using history is a way to manage partial observability, we tried it. Figure 4 shows the results of using G in the two domains. While the deictic agents certainly learn faster than any of the agents learned using NDP, the deictic agents with G never learn the task as *well* as the full-propositional agent does with NDP. Furthermore, the full-propositional agent was never able to get off the ground with G.

## 5 Discussion

Because the goal of our work was to understand the characteristics of these learning approaches, rather than to build a particular working demonstration, we continued with a program of experimentation aimed at elucidating our counter-intuitive results.

### 5.1 On Deictic and Full-Propositional Representations in NDP

The optimal policy for the deictic agent is to start with a *focus-on(green)* action, then to move the focus up (until the top of the stack is reached), then to pick up the top block and move it to the side. This sequence should repeat until the green block is uncovered and picked up. In both blocks-world setups, this requires a sequence of nine actions. In the full-propositional case, the optimal policy is tedious but generates shorter action sequences. It goes roughly as follows: if *block-1* is green and clear, then the pick up *block-1*; otherwise, if

---

[3]For each of the three representations, the reward total was scaled by the maximum reward achievable by the optimal policy.

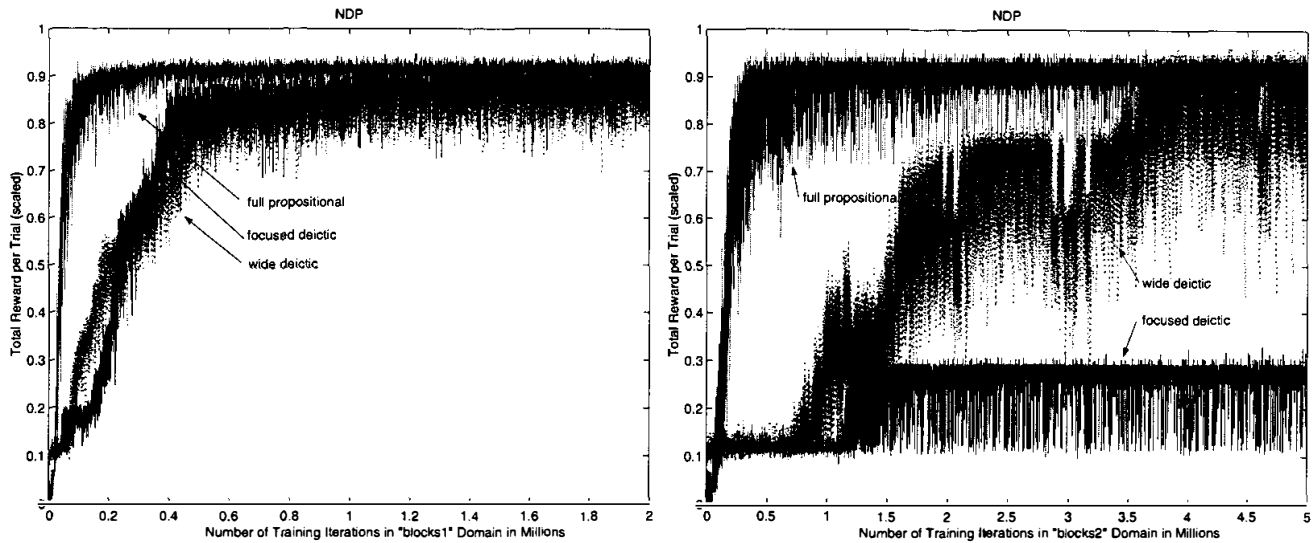

Figure 3: Learning curves for NDP in a) the *blocks1* domain and b) the *blocks2* domain.

*block-2* is green and clear, then pick up *block-2*; etc. If there is no block that is green and clear, then if *block-1* is on top of a green block and is clear, pick up *block-1*; etc. In both blocks-world setups, the optimal policy requires a sequence of four actions. While the required action sequence is short, the same ideas have to be represented over and over for each assignment of names to blocks.

We initially reasoned that the deictic agent had more trouble learning an optimal policy because it appeared to have a harder exploration problem. The dependence of the *pick-up()*, *marker-to-focus(marker)*, and *focus-to-marker(marker)* actions on the focus location means that it is very easy for the agent to lose its place by executing an exploratory action that moves the focus. The result is that the outcome of the above actions can be wildly different depending on where the focus happened to land. This "distractability" reduces the effectiveness of using exploration to make learning progress.

In analyzing the causes of the exploration problem, we did experiments that ruled out the longer optimal action sequence and the number of actions in the action set [6]. Our final step was to created a modified action set. In this new action set, the *pick-up()* action automatically picks up the block at the top of the stack pointed to by the focus, and the *marker-to-focus(marker)* and *focus-to-marker(marker)* actions were removed.[4] Otherwise, the action set was the same as the original action set. The implication of changing the *pick-up()* action in this way is that the action is now more likely to result in a successful pickup, since the agent cannot even try to pick up blocks that are not clear (i.e., the blocks in the middle of a tall stack). In other words, the outcome of the all-important *pick-up()* action loses its absolute dependence on the focus location, resulting in more robustness in the face of a moved focus. As we shall see, the modified action set rendered exploration much more effective in pointing the agent towards the goal.

To compare the effects on exploration of the original and modified action sets, we measured, for each representation, the number of steps required by a random agent to stumble upon a solution. This metric, the mean time-to-goal, is plotted as a function of the number of distractor blocks in Figure 5. It is clear from the figure that the modified deictic action set makes it much easier to achieve the goal via a random walk; with the modified actions, exploration in the deictic system scales in the same way as in the propositional system. Follow-up learning experiments with the modified action set in NDP show the deictic agents on average learn as fast the full-propositional agent in *blocks1* and slightly faster than the full-propositional agent in *blocks2* [6].

However, it is important to note that, in general, neither the full-propositional nor the deictic agents with the modified action set *learn* an optimal policy here. The full-propositional agent does not because it must learn a policy for each way to name the blocks and this takes a long time; the deictic agents do not because an optimal policy for this action set still has a crucial dependence on past history. That is, the behavior of the deictic agent's actions is dependent on the focus location; yet, the none of the marker locations are observable by the agent—it must be recovered by examining past history. An unfortunate exploratory choice causes problems because the history that must be used to recover the focus location now includes the exploratory action. The reason the modified action set leads to better learning is because a plausible policy is available that does not rely on history at all.[5]

---

[4]Interestingly, this modified action set is similar to the set used by McCallum in his blocks-world experiments [9].

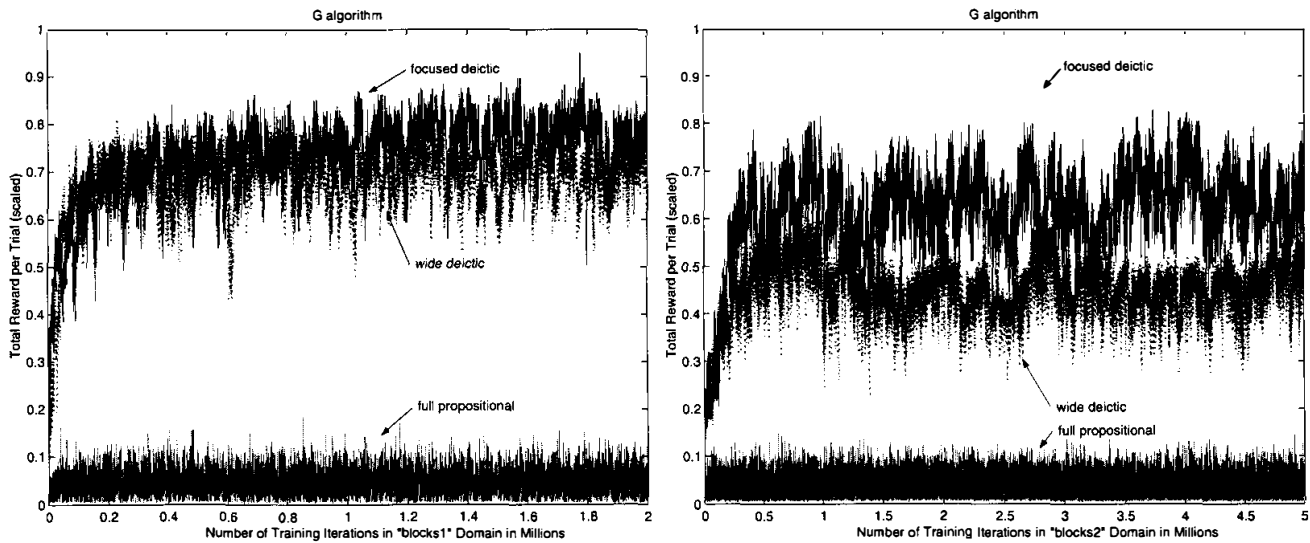

Figure 4: Learning curves for G algorithm in both domains.

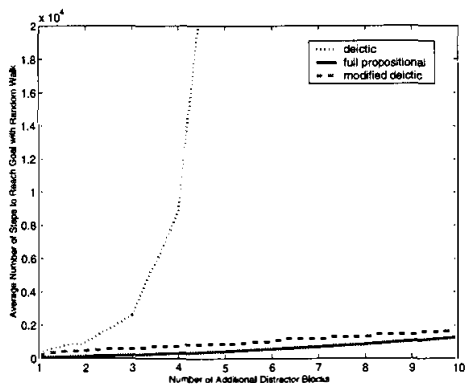

Figure 5: The mean time-to-goal for different action sets plotted against the number of distractor blocks.

The deictic state and action sets not only fail to exhibit the advantages we expected to find, but introduce new challenges to learning that must be overcome in order to make effective use of such representations. We conclude that it is possible to tailor the action set to the task so that a deictic representation is more feasible, but the flexibility of such an action set is obviously more limited. An action set that includes the ability for the agent to control its own attentional focus inherently increases the difficulty of the exploration problem because of the information about the domain stored implicitly in the location of the focus (and potentially the markers). Thus, any exploratory actions that move the focus at random make it very hard for the agent to learn a useful policy. McCallum [9], in his blocks-world task with the more carefully-tuned action representation, also found that learning was much improved when the exploration was guided by a human. The implications of the dependence of the optimal policy on history is examined in the next section.

---

[5]Namely: look at the green block; pick up the block at the top of the stack (if the green block was clear, this leads to the goal); if my hand is full, look at the table; if I'm looking at the table, put the block down at the top of this stack.

### 5.2  On NDP and G

The common wisdom is that function approximators like neural nets are appropriate for problems in which all of the input attributes are relevant to some small degree, and that decision trees are appropriate when the function is well represented in terms of an unknown subset of the input features. In the full-propositional representation, any of the bits could be important, so it seems reasonable that NDP worked well. In the deictic representation, we included many historical observations into the input vector, not knowing which ones might be relevant to the problem. In this situation, we might expect the tree-growing algorithms to be better: they should build a representation that reveals only enough of the hidden state to do the job.

Our poor results with the G algorithm seem to be primarily caused by the trees growing much larger than expected; they grew without reaching a natural limiting state. To avoid running out of memory, we had to add an arbitrary cap on the size of the trees.

While the deictic agents initially learn faster with G than with NDP, they stopped making progress upon reaching the tree-size cap, and therefore never completely learn the task. Similarly, the full-propositional agent made no progress at all before the tree reached its maximum size. In the process of determining the cause behind the trees' seemingly unnecessarily large size, we discovered the true root of the problem that was preventing our deictic agents from learning effectively.

### 5.2.1 Unlimited Tree Growth

Given the ability to characterize the current state in terms of past actions and observations, the learning algorithm frequently comes up with multiple perceptual characterizations that correspond to the same underlying world state. For instance, the set of states described by *the focus was on a green block and then I looked up* is the same as those described by *the focus was on a green block and then I looked down, then up, then up*, etc. In isolation, these redundant leaves do not seem to pose much of a problem—one solution would be an algorithm that grows a graph rather than a tree, allowing for information states that represent the same underlying state to be merged. However, in a tree, the impact of these redundant nodes can be severe.

Once the tree contains multiple leaves for the same state based on historical distinctions, the agent may now learn to have a different policy in this state, depending on its previous actions. This complicated policy now requires still more splits to fully learn the value function. There is, fundamentally, a kind of "arms race," in which a complex tree is required to adequately explain the Q values of the current policy. But the new complex tree allows an even more complex policy to be represented, creating a vicious cycle. The basic problem is that the tree is trying to grow enough leaves to learn the value function for *every* policy followed during the learning process.

### 5.2.2 Incurable Partial Observability

In investigating the problem of the very large trees, we attempted to build a tree by hand that would allow the agent to learn the optimal policy while containing as few leaves as possible. What we discovered instead was that it is not possible for any amount of history to make the domain Markov. This is because not all history sequences allow the agent to disambiguate otherwise identical-looking states. Consider the example given at the beginning of Section 2, where the agent is looking at the red block. Clearly the two history sequences described in that section allow the agent to determine whether it has already cleared the green block or not. However, if the agent was looking at the table, performed a focus_color(red), and is now looking at the red block, we have no idea which of these two possible states we are in. Thus, the tree will always contain leaves that are ambiguous with respect to the true underlying state, and no amount of further splitting will remedy this.

Thus, no matter how large the trees get, the agent is still trying to learn in a partially observable domain. In fact, this problem is present regardless of which learning algorithm is used; the problem is simply not Markovian, no matter how much history is added to the agent's observation.

## 6 Conclusion

In the end, none of the approaches for converting an inherently relational problem into a propositional one seems like it can be successful in the long run. The naïve propositionalization grows exponentially in the number of objects in the environment; even worse, it is severely redundant due to the arbitrariness of assignment of names to objects. The deictic approach has a seemingly fatal flaw: the inherent dramatic partial observability poses problems for model-free value-based reinforcement learning algorithms. The fundamental problem with using short-term history in a POMDP is this: the ability to disambiguate underlying states is necessary for learning a good policy, but past actions and observations are not useful data *until* a good policy is available.

One possible direction to consider before abandoning this approach altogether would be to adopt Perkins' provably convergent algorithm for Monte Carlo learning in a partially observable domain [13]. While this algorithm is known to converge, it is not yet known whether it will converge to a desirable answer in this case.

Alternatively, one could change the approach more fundamentally. There are three strategies to consider, two of which work with the deictic propositional representation but forgo direct, value-based reinforcement learning.

One alternative to value-based learning is direct policy search [17, 7], which is less affected by problems of partial observability but inherits all the problems that come with local search. It has been applied to learning policies that are expressed as stochastic finite-state controllers [11], which might work well in the blocks-world domain. These methods are appropriate when the parametric form of the policy is reasonably well-known *a priori*, but probably do not scale to very large, open-ended environments.

Another strategy is to apply the POMDP framework more directly and learn a model of the world dynamics that includes the evolution of the hidden state. Then, we might use reinforcement-learning algorithms to more successfully learn to map this mental state to actions.

A more drastic approach is to give up on propositional representations (though we might well want to use deictic expressions for naming individual objects), and use real relational representations for learning in blocks world. Some important early work has been done in relational reinforcement learning [5], showing that relational representations can be used to get appropriate generalization in complex completely observable environments.

### Acknowledgments

This work was funded by the Office of Naval Research contract N00014-00-1-0298, by the Nippon Telegraph & Telephone Corporation as part of the NTT/MIT Collaboration Agreement, and by a National Science Foundation Graduate Research Fellowship.